\documentclass{IEEEtran} 

\usepackage{array}
\usepackage{graphicx}%
\usepackage[compatibility=false]{caption}
\usepackage{subcaption}
\usepackage{multirow}%
\usepackage{amsmath,amssymb,amsfonts}%
\usepackage{amsthm}%
\usepackage{mathrsfs}%
\usepackage{bm}%
\usepackage{xcolor}%
\usepackage{textcomp}%
\usepackage{manyfoot}%
\usepackage{booktabs}%
\usepackage{algorithm}%
\usepackage{algorithmicx}%
\usepackage{algpseudocode}%
\usepackage{listings}%
\usepackage{enumitem}
\usepackage{placeins}
\usepackage{rotating}
\usepackage{lineno}
\usepackage{diagbox}
\usepackage{slashbox}
\usepackage[hyphens,spaces,obeyspaces]{url}
\usepackage[breaklinks=true]{hyperref}
\usepackage{makecell}
\newcolumntype{L}[1]{>{\raggedright\arraybackslash}p{#1}}

\usepackage[final]{changes}

\title{\LARGE \bf News and Load: Social and Economic Drivers of Regional Multi-horizon Electricity Demand Forecasting}
\author{
\IEEEauthorblockN{Yun Bai\IEEEauthorrefmark{1}, \textit{Member, IEEE},
Simon Camal\IEEEauthorrefmark{2}, \textit{Member, IEEE},
Andrea Michiorri\IEEEauthorrefmark{2}}
\thanks{\IEEEauthorrefmark{1}Corresponding author.}
\thanks{\IEEEauthorrefmark{2}These authors contributed equally to this work.}  
\thanks{The authors are with the Centre for Processes, Renewable Energies and Energy Systems (PERSEE), MINES Paris - PSL University, Sophia Antipolis, France, e-mail: (name.surname@minesparis.psl.eu). The author Yun Bai was supported by the program of the China Scholarship Council (CSC Nos. 202106020064). Part of this work is carried out in the framework of the Fine4CAST project, funded by France 2030 (ANR reference: 22-PETA-0008).}%
}

\begin{document}

\maketitle
\thispagestyle{empty}
\pagestyle{empty}

\begin{abstract}
The relationship between electricity demand and variables such as economic activity and weather patterns is well established.
However, this paper explores the connection between electricity demand and social aspects. 
It further embeds dynamic information about the state of society into energy demand modelling and forecasting approaches.
Through the use of natural language processing on a large news corpus, we highlight this important link.
This study is conducted in five regions of the UK and Ireland and considers multiple time horizons from 1 to 30 days. 
It also considers economic variables such as GDP, unemployment and inflation.
The textual features used in this study represent central constructs from the word frequencies, topics, word embeddings extracted from the news.
The findings indicate that: 1) the textual features are related to various contents, such as military conflicts, transportation, the global pandemic, regional economics, and the international energy market.
They exhibit causal relationships with regional electricity demand, which are validated using Granger causality and Double Machine Learning methods.
2) Economic indicators play a more important role in the East Midlands and Northern Ireland, while social indicators are more influential in the West Midlands and the South West of England.
3) The use of these factors improves deterministic forecasting by around 6\%.
\end{abstract}



\section{Introduction}\label{Introduction}

Electricity demand forecasting is linked to various factors, including weather, economic activity, and major events.
These factors have been extensively studied across different spatial and temporal scales \cite{gerossier2018robust}. 
First, weather conditions significantly affect electricity demand by changing human activities, such as the use of heating, cooling, and lighting devices \cite{macmackin2019modeling}.
Second, the intensity and development level of economic activity are key elements influencing electricity demand.
For example, business hours and weekdays are characterised by higher electricity demand \cite{li2020midterm}. 
Macroeconomic indicators such as Gross Domestic Product (GDP) per capita have been found to influence national electricity demand significantly \cite{bianco2009electricity}.
Third, large events lead to changes in consumption patterns both temporally and spatially.
For example, the lockdown policies during COVID-19 reduced electricity usage \cite{ruan2020cross, farrokhabadi2022day}.

In the early 2000s, researchers pointed out that social sciences should be integrated into energy demand studies in order to gain a comprehensive understanding of the formation and evolution of energy demand, which is influenced by market prices, consumer awareness, and social norms \cite{wilhite2000legacy}.
However, social factors pose challenges due to their ambiguous definition and measurement.
In social sciences, scientists depict the complex relationship between human social characters and energy demand through household interviews, surveys, and relational sociology. 
These studies offer insights into human behaviour and its correlation with peak demand, flexibility volumes, and energy usage patterns \cite{powells2014peak,wagy2017crowdsourcing,hargreaves2020importance}.
More recently, studies have been conducted to better model electricity demand with large datasets related to social activities, including mobile phone network data \cite{doumeche2023human}, online search queries \cite{fu2022using}, and high-resolution satellite images \cite{allen2016impacts}.

Public news contains extensive but overlooked social knowledge that is potentially valuable for electricity demand modelling. 
Usually, the information is noisy, unstructured, and sparse, making it challenging to analyse and interpret \cite{subramaniam2009survey}.
The emergence of Natural Language Processing (NLP) technologies allows us to discover knowledge from massive text.
Attempts have already been made to forecast energy by incorporating news text, such as crude oil prices \cite{li2019text, wu2021effective, liu2021forecasting}.
In electricity demand forecasting, the study by \cite{obst2021adaptive} used key word frequency extracted from social media to improve forecasting and explain the structural change in the load profiles.
This research is an inspiring attempt at text-based electricity demand forecasting, but the incorporation of richer text features remains an issue.
Other studies have worked on latent features processed by deep networks, for example Convolutional Neural Network (CNN) and Long Short-Term Memories (LSTM) \cite{wang2023forecasting, bai2023electricity}.
Although they provided improved forecasts compared to benchmarks, these studies lacked explanation of how the textual features worked, which is more compelling in this research.
In \cite{bai2024newsandload}, the authors discovered several social variables in unstructured news that could cause changes in electricity demand.
However, the use cases only concern UK and Ireland national day-ahead load forecasting.
Regional scenarios and longer forecasting terms should be explored to verify the time-varying impacts of social factors on electricity demand.


Based on the state-of-the-art, there are still research gaps in text-based electricity forecasting: i) the improved performance should be explained to avoid the effects of spurious correlations; 
ii) recent studies of demand forecasting with text have focused on single time series or forecast horizon, overlooking the potential for generalising the results and preventing the opportunity to determine the most appropriate application.
iii) the predominance of deterministic forecasting prevents a comprehensive analysis of the ability of the method to deal with forecasting uncertainty.

In this study, we propose a methodology to analyse the relationship between news and electricity demands from multiple perspectives.
We evaluate the method on a case study of five regions in the UK and Ireland, and forecasting horizons from 1 to 30 days ahead. This study allows readers to understand where and when the social indicators are more relevant.
This study answers the following five research questions of relevance to power system practitioners, with more detailed discussion provided in Section~\ref{Discussion}:
\begin{enumerate}
    \item Is there any relationship between national news and regional electricity demand?
    \item Can textual news be used for practical applications to improve regional demand forecasting?
    \item Which topics have a higher impact on electricity demand?
    \item Are the results consistent for different regions?
    \item For which forecast horizons is the impact higher?
\end{enumerate}

In the rest of the paper, Section~\ref{Methodology} introduces the methods used in this study. 
Section~\ref{Results} presents the results of both deterministic and probabilistic forecasting, and the analysis of causality and important features.
Section~\ref{Discussion} discusses the research questions above, and Section~\ref{Conclusions} concludes the paper.
In order to facilitate the application to the wider research community, the codes and datasets are made available\footnote{\href{https://github.com/YunBAI-PSL/Regional-electricity-demand-forecasting-with-news}{https://github.com/YunBAI-PSL/Regional-electricity-demand-forecasting\linebreak-with-news}}.

\section{Methodology}\label{Methodology}
This section presents both the engineering of numeric features and textual features in Subsection~\ref{Feature engineering}, describes the model configuration in Subsection~\ref{Model Configurations}, and introduces the methods for causality analysis in Subsection~\ref{Model explanation}.

\subsection{Feature Engineering}
\label{Feature engineering}
\subsubsection{Numerical Features}
The benchmark model contains 65 numerical features of historical lags, calendar information, and temperatures.
The demand for 24 hours on the target day $d+h$, $h$ horizons ahead of the issued day $d$, is considered.
In calendar features, there is a 0-1 variable to show if the target day is a weekday or weekend, 12 dummy variables of bank holidays or normal day indicators, and the sine and cosine waves of day-of-week and day-of-year are included to capture weekly and yearly patterns.
Given that longer forecasting horizons, the averaged hourly and monthly temperatures were employed for the target day, instead of observations, to maintain the central trend and seasonality, resulting in a total of 24 variables.

\subsubsection{Textual Features}\label{Methods:Textual features}
\begin{figure}[ht]
\centering
\includegraphics[scale=.35]{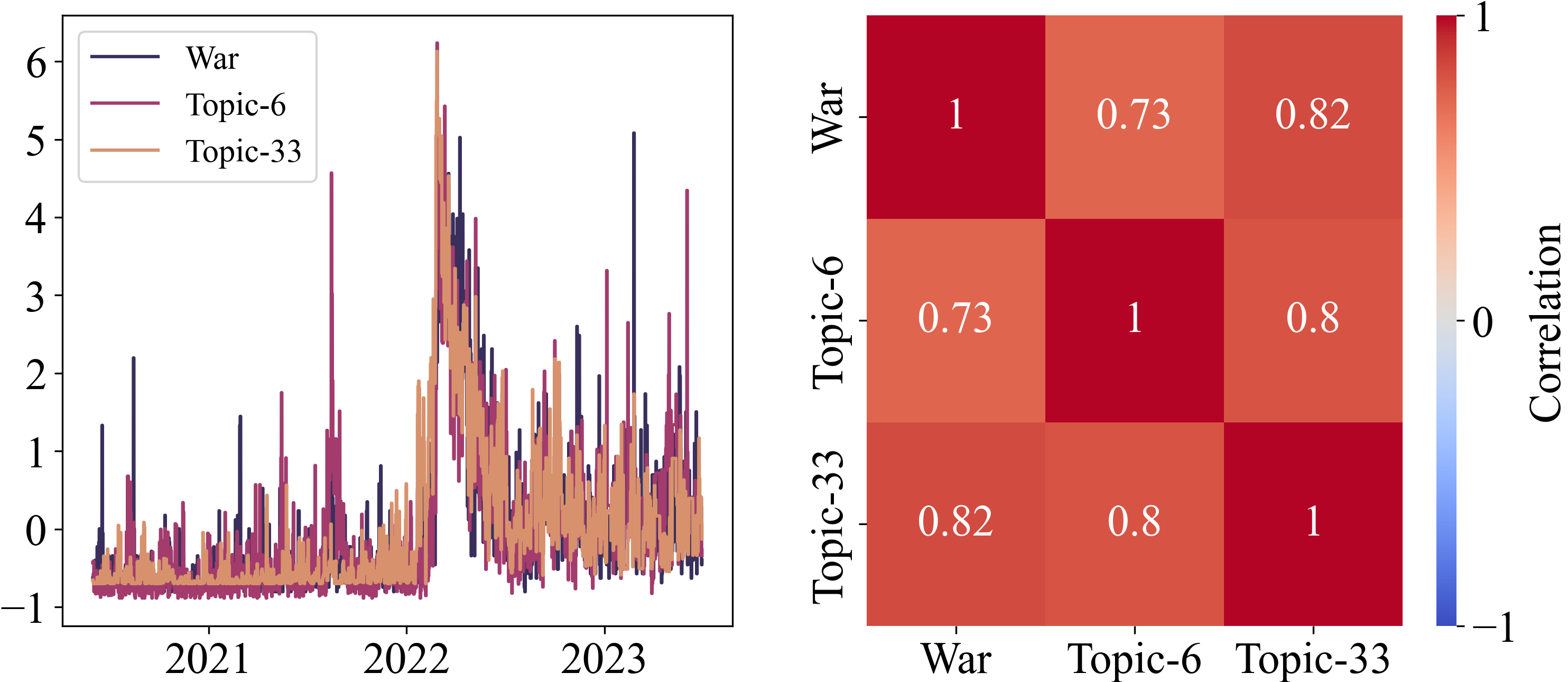}
\caption{\textbf{(a)} Time series, and 
\textbf{(b)} Pearson correlations between war-related features. The three variables have been normalised.}
\label{war-related features}
\end{figure}

In the work \cite{bai2024newsandload}, the authors extracted thousands of features from a vast corpus of news texts using various NLP methods, including statistical counts, word frequencies, public sentiments, topic distributions, and word embeddings.
The authors noted that not all the features are useful for forecasting, and features from different sources may overlap semantically. 
For example, the frequency of the word `war' and war-related topic features refer to the same content, and their feature series have similar shapes and are highly correlated.
Figure~\ref{war-related features} presents the time series of word frequency of `war' and two war-related topics, along with the Pearson correlation between the variables.
The keywords of Topic-6 are: city, forces, war, troops, Kyiv, military, killed, region, soldiers, Zelensky, civilians, fighting, capital, east, attacks.
The keywords of Topic-33 are: Russian, Russia, Ukraine, Ukrainian, Putin, president, war, Moscow, invasion, NATO (North Atlantic Treaty Organization), Vladimir, military, sanctions, Russians, western.
Although we can discern the subtle differences of the two topics that Topic-6 describes the battlefield situation and Topic-33 focuses on the reaction of western countries, they are both about the Russia-Ukraine War.

The semantic overlap among multiple features can introduce redundancy and noise into forecasting models, making it essential to simplify the feature space. 
To address this, we developed an efficient feature aggregation approach using clustering, allowing the model to better utilise the streamlined features and avoid overemphasis on specific or overlapping ones.
The proposed method has three advantages: i) avoiding working with a large number of features by focusing on only ten centroids; ii) reducing the semantic overlaps between features by maximising the distances among clusters; iii) maintaining interpretability while simplifying the analysis procedure and reducing computing time.

\begin{figure}[ht]
\centering
\includegraphics[scale=.5]{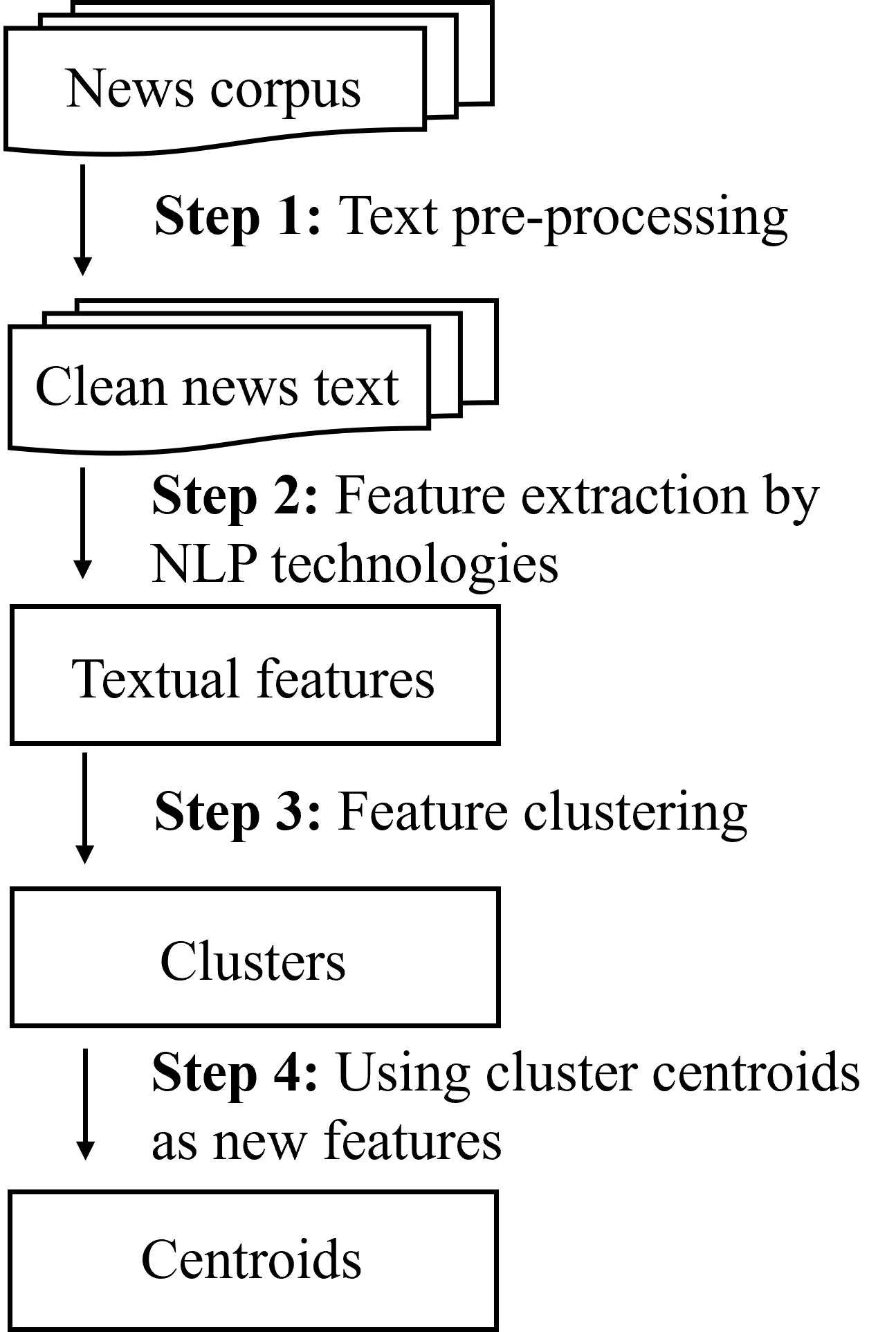}
\caption{Workflow of textual feature extraction process.}
\label{Workflow of textual feature extraction process.}
\end{figure}

The workflow is outlined in Figure~\ref{Workflow of textual feature extraction process.}, where the first two steps were completed in \cite{bai2024newsandload}. 
Step 3 involves grouping textual features into clusters using hierarchical clustering. 
Compared to K-Means, the hierarchical method does not rely on random initialisation and produces stable results, which are organised into a dendrogram. 
The dendrogram provides a visual representation of the global structure, allowing users to determine the number of clusters by applying Ward's Minimum Variance Method \cite{ward1963hierarchical}.
The Ward method minimises the total within-cluster variance by iteratively merging clusters to minimise information loss at each step. 
The height of the dendrogram at each merge reflects the Ward value, which indicates the variance increase caused by merging, and this can guide the selection of the optimal number of clusters. 
As a result, there is no need to predefine the number of clusters $k$ or re-run the clustering model for each $k$, as required in K-Means.

\begin{figure}[ht]
\centering
\includegraphics[scale=.45]{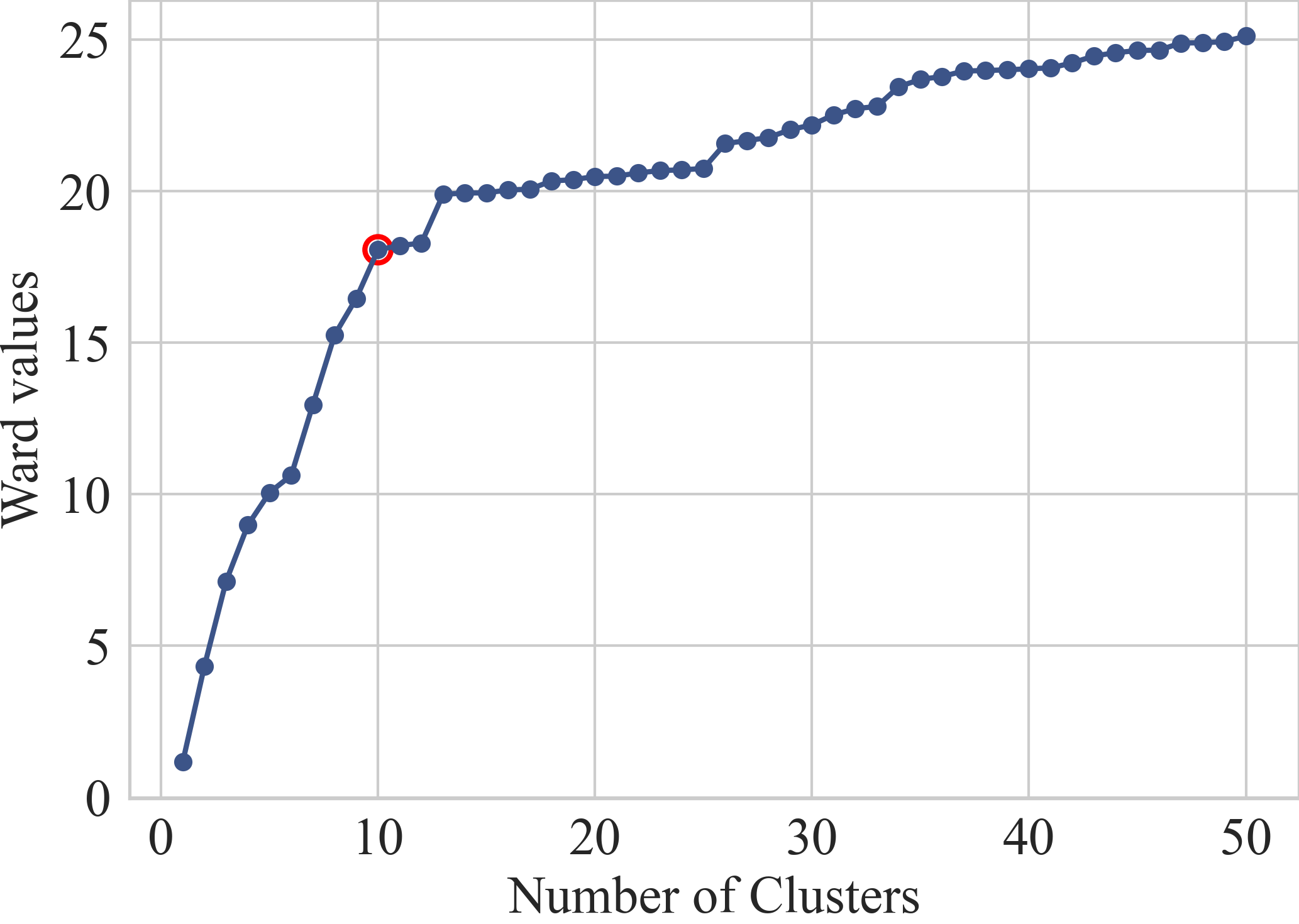}
\caption{Ward values for the selection of the number of cluster. The elbow point corresponds to 10 clusters.}
\label{elbow_cluster_select}
\end{figure}

Figure~\ref{elbow_cluster_select} illustrates the Ward values against the number of clusters. 
The elbow is clearly observed at 10 clusters (marked in red), beyond which the rate of increase in Ward values significantly slows down. 
This suggests that 10 clusters effectively balance the trade-off between reducing dimensionality and preserving feature information.

\begin{figure*}[ht]
\centering
\includegraphics[scale=.31]{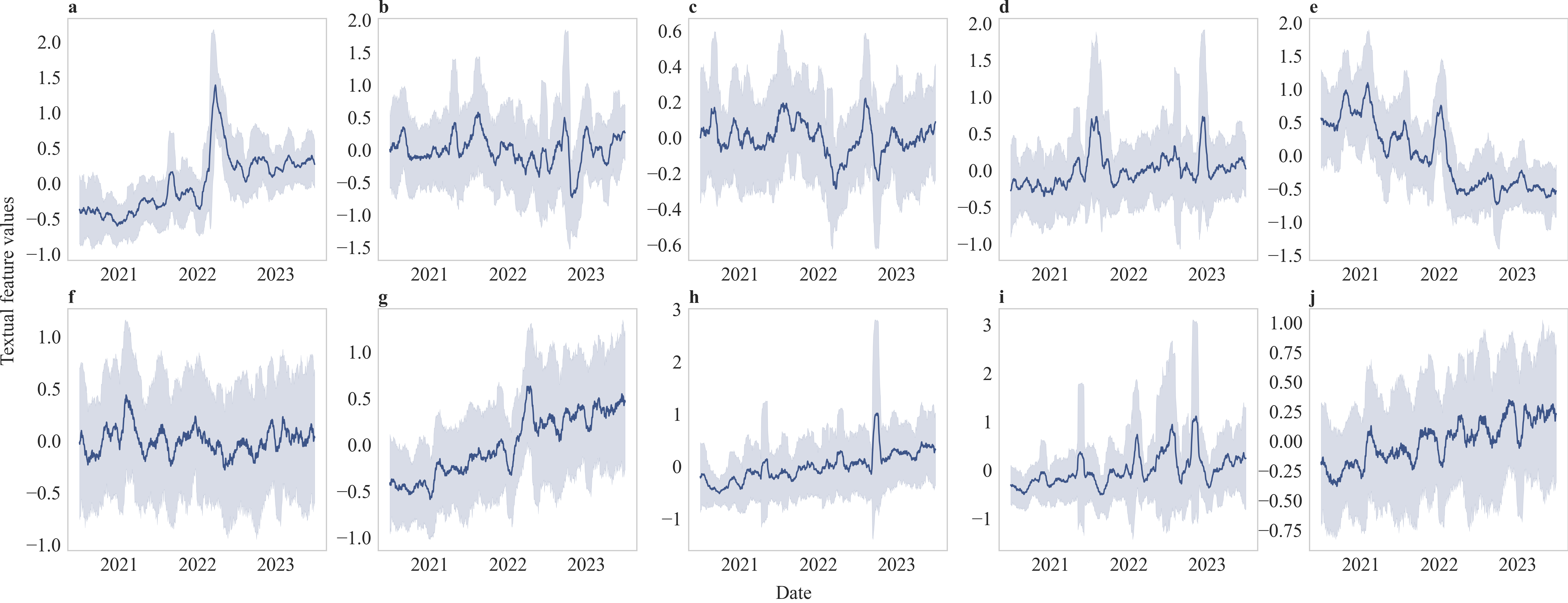}
\caption{Clustering centroids of the textual features: \textbf{a:} military conflicts, \textbf{b:} transportation, \textbf{c:} travel \& leisure, \textbf{d:} sports events, \textbf{e:} pandemic control, \textbf{f:} regional economics, \textbf{g:} strikes, \textbf{h:} family life, \textbf{i:} election, \textbf{j:} energy markets.}
\label{TextualCentroids}
\end{figure*}

In Step 4 of Figure~\ref{Workflow of textual feature extraction process.}, we took the centroids from each cluster as social factors and summarised the content of them according to the topics and keywords within the cluster.
The ten centroids are plotted as shown in Figure~\ref{TextualCentroids}.
In \textbf{(a)} military conflicts, there is noticeable peak related to the Russia-Ukraine War.
In \textbf{(e)} pandemic control, decreasing trend can be observed due to the reducing COVID-19 cases.
An increasing trend can be found in feature strikes \textbf{(g)} and energy markets \textbf{(j)}.

In fact, the entire textual analysis process of Figure~\ref{Workflow of textual feature extraction process.} can be regarded as a sequence of information refinement steps. 
In \textbf{Step 1}, irrelevant elements such as stop words and punctuation marks are filtered out. 
In \textbf{Step 2}, specific semantic features, such as topics and sentiments, are extracted. 
While advanced NLP techniques could capture deeper linguistic information, we focus on retaining meaningful textual features with interpretability.
During \textbf{Step 3}, feature series are treated as sample points for clustering. 
Some information, such as fine-grained feature distinctions, is inevitably lost. 
However, this loss is minimised as the clustering process focuses on grouping similar series, ensuring that the most relevant patterns are preserved.
Finally, selecting cluster centroids in \textbf{Step 4} discards some intra-cluster information, but this trade-off reduces noise in forecasting models. The centroids effectively summarise key information and simplify the feature space.

\subsection{Model Configurations and Evaluation Metrics}
\label{Model Configurations}

This study considers deterministic and probabilistic forecasting for regional electricity demand. 
We focus on Gradient Boosting Machines (GBM) for both tasks, as GBM have the flexibility to incorporate textual features. 
Specifically, Light GBM (LGBM) and Probabilistic GBM (PGBM) are employed for deterministic and probabilistic forecasting, respectively.

Before model training, the dataset is split into training, validation, and test sets.
All the data are standardised based on the training set to ensure that the features are on the same scale.
For multi-horizon forecasting tasks for each region, we built 30 models for 30 horizons.
In each horizon, we wrapped the hourly data into a multi-regression form, with 24 targets for 24 hours one day ahead.
For sample $i$ and horizon $h$,  the formulas are $\hat{\bm{y}}_{i,h} = f_d(\bm{x}_i)$ for deterministic forecasting and $(\bm{\mu}_{\hat{y}_{i,h}}, \bm{\sigma}_{\hat{y}_{i,h}}^2) = f_p(\bm{x}_i)$ for probabilistic forecasting, where $\bm{x}_i=(x_{i,1},x_{i,2},...,x_{i,m})$ is the vector with $m$ features, $f_d(*)$ and $f_p(*)$ are LGBM and PGBM models, $\hat{\bm{y}}_{i,h} = (\hat{y}_{1,i,h},\hat{y}_{2,i,h},...,\hat{y}_{24,i,h})$ is the target vector with 24 hours, $\bm{\mu}_{\hat{y}_{i,h}} = (\mu_{\hat{y}_{1,i,h}},\mu_{\hat{y}_{2,i,h}},...,\mu_{\hat{y}_{24,i,h}})$ and $\bm{\sigma}_{\hat{y}_{i,h}}^2 = (\sigma_{\hat{y}_{1,i,h}}^2,\sigma_{\hat{y}_{2,i,h}}^2,...,\sigma_{\hat{y}_{24,i,h}}^2)$ are mean and variance vectors with 24 hours.
$f_d(*)$ and $f_p(*)$ both use Mean Squared Error (MSE) on the validation set as the loss function. 
For hyper-parameter tuning, we employed Optuna, an automatic optimisation framework \cite{akiba2019optuna}.
Optuna designs efficient searching and pruning strategies and allows users to build search space dynamically, thus achieving high performance with limited computational resources.
For simplicity, we refer to both LGBM and PGBM as GBM in the main text.

For deterministic forecasting, Root Mean Square Error (RMSE) and Mean Absolute Percentage Error (MAPE) are used to measure the deviation between the actual values and the forecasts.
Continuously Ranked Probability Score (CRPS) measures the discrepancy between the cumulative distributions of the truth and forecasts.
The three metrics can be calculated at a given horizon as follows:
\begin{equation}
    \text{RMSE} = \sum_{d=1}^D \sqrt{\frac{1}{24}\sum_{h=1}^{24}(y_{h,d} - \hat{y}_{h,d})^2},
\end{equation}
\begin{equation}
    \text{MAPE} = \frac{100\%}{D\times 24}\sum_{d=1}^D \sum_{h=1}^{24} \left| \frac{y_{h,d} - \hat{y}_{h,d}}{y_{h,d}} \right|,
\end{equation}
\begin{equation}
\label{CRPS_formula}
    \text{CRPS} = \frac{100\%}{D\times 24}\sum_{d=1}^D \sum_{h=1}^{24} \int [F(\hat{y}_{h,d}) - \mathbf{1}_{\{\hat{y}_{h,d} \geq y_{h,d}\}}]^2d\hat{y},
\end{equation}
where RMSE and MAPE are first calculated for each day ahead (24 hours $h$), then averaged over all of the days $D$ in test set.
$y_{h,d}$ and $\hat{y}_{h,d}$ are the truth and forecasts.
In Eq~(\ref{CRPS_formula}), CRPS is calculated on each hour and averaged similarly as RMSE and MAPE.
$F(\hat{y}_{h,d})$ is the Cumulative Distribution Function (CDF) of $\hat{y}_{h,d}$.
$\mathbf{1}$ is the indicator function.

\subsection{Causality Analysis}
\label{Model explanation}

\subsubsection{Granger Test}
Based on the Vector Auto-Regression (VAR) model, the purpose of the Granger test is to detect whether a time series $X$ can forecast another series $Y$ \cite{granger1969investigating}.
If so, it can be said that series $X$ Granger-causes $Y$.
To conduct the test, two VAR models are constructed for $Y$:
\begin{equation}
    Y_t = \alpha_0 + \alpha_1 Y_{t-1} + \cdots + \alpha_p Y_{t-p} + \varepsilon_t,
\end{equation}
\begin{equation}
    Y_t = \beta_0 + \beta_1 Y_{t-1} + \cdots + \beta_p Y_{t-p} + \gamma_1 X_{t-1} + \cdots + \gamma_q X_{t-q} + \epsilon_t,
\end{equation}
where $p$ and $q$ are lag orders, $\alpha_i$, $\beta_i$, and $\gamma_i$ are coefficients, $\varepsilon_t$ and $\epsilon_t$ are error terms.
By comparing these models, the Granger test determines whether lagged terms of $X$ improve the prediction of $Y$. 
If the second model significantly outperforms the first, it is concluded that $X$ Granger-causes $Y$.

\subsubsection{Double Machine Learning (DML)}

DML is another framework for causal inference that isolates the causal effect of a treatment $T$ on an outcome $Y$ while controlling for high-dimensional covariates $X$ \cite{chernozhukov2018double}. 
The method involves three key steps: first, a model estimates $\hat{T}=f_1(X)$, measuring the influence of $X$ on $T$; 
second, another model estimates $\hat{Y}=f_2(X)$, capturing the relationship between $Y$ and $X$. 
Residuals are then calculated as $T-\hat{T}$ and $Y-\hat{Y}$ to remove the confounding effects of $X$, and the causal effect $\delta$ is estimated by fitting an ordinary least squares regression on the residualised variables:

\begin{equation}
    Y-\hat{Y} = \delta(T-\hat{T}) + \epsilon,
\end{equation}
where $\epsilon$ is the error term.

The Granger test focuses on detecting predictive relationships. 
However, it does not always reflect true causality, as it captures statistical dependencies. 
In contrast, DML isolates the causal effect of $T$ on $Y$, enabling a more granular analysis. 
In this study, the Granger test is used to examine the overall relationship between textual features and regional electricity demand, and DML provides a detailed view of causal variations under different horizons. 
This combination enables the exploration of both macro-level predictability and fine-grained causal effects.

\section{Results}\label{Results}

This section first shows the datasets used in this study.
Considering both deterministic and probabilistic forecasting, this section compares the error metrics of the models with and without economic and social factors across different regions.
We also illustrates an example of forecasting on a bank holiday.
To understand the mechanism of how economic and social factors impact forecasting, the last two parts analyses feature importance and causation of social factors and demand. 

\subsection{Dataset}\label{Results: Dataset}
As an extended study of \cite{bai2024newsandload}, this work selects the UK and Ireland as the research cases.
The regions in the UK are East Midlands, West Midlands, South Wales, South West, and Northern Ireland.
Historical electricity demand is obtained from \cite{NationalGridWeb} and the ENTSO-E transparency platform \cite{entso-e}.

News articles are sourced from the BBC news repository, which is maintained by \cite{BBC_news}.
Textual features such as word frequencies and topics are extracted by \cite{bai2024newsandload}.
The thousands of features generated are clustered and aggregated into social factors as described in Section~\ref{Methods:Textual features}.
We selected BBC news for its authority and extensive coverage, collecting 57,686 articles over three years.
Previous research shows that high-quality news sources provide broader information and stronger predictive power \cite{rambaccussing2020forecasting}.
BBC News has already been validated for national-level electricity forecasting \cite{bai2024newsandload}, and future work will explore additional sources to enhance regional predictions.

We consider economic data for the following reasons: i) to ensure that any causality observed between social factors and electricity load is genuine; 
ii) to ensure that performance improvements are not just due to easily available economic data and that social factors provide additional value rather than simply reflecting economic trends.
We apply GDP, inflation, and unemployment rate from a national scale to incorporate the macroeconomic effects on electricity demand from \cite{economic_factor}.

Unlike \cite{bai2024newsandload}, with regard to regional temperatures, 2-meter temperatures for the major cities in each region were obtained from ERA5, the global climate and weather dataset from the European Centre for Medium-Range Weather Forecasts (ECMWF) \cite{hersbach2023era5}.
Concretely, we collected temperatures from Leicester for the East Midlands, Birmingham for the West Midlands, Cardiff for South Wales, Bristol for the South West, Belfast for Northern Ireland, and Dublin for Ireland.

After performing feature engineering as described in Section~\ref{Feature engineering}, we divided the data into a training set from 2020-06-01 to 2022-11-30, a validation set from 2022-12-01 to 2022-12-31, and a test set from 2023-01-01 to 2023-05-31.

\subsection{Deterministic Forecast} \label{Results: Deterministic forecast}

This subsection takes the LGBM model for incorporating economic and social factors, as it performs well across all the candidate benchmarks, as shown in Appendix~\ref{Forecasting Results of Benchmark Models}.
To compare the individual and combined impacts of these factors on electricity demand, we designed an experiment involving four models trained with different data sets.
The four models are: GBM, trained only with numerical features; GBM-E, which adds economic indicators (GDP, inflation, and unemployment) to GBM; GBM-S, which incorporates social factors instead; and GBM-ES, which combines both economic and social variables.
Table~\ref{Model performance table of RMSE} shows the RMSE and MAPE performance across all the regions on the average of 30 horizons.

\begin{table}[ht]
\centering
\caption{Horizon-averaged model performance of RMSE and MAPE with or without economic and social factors. S and E indicate social and economic factors. N (No) and Y (Yes) indicate whether the factors (S or E) are included in the model.}
\label{Model performance table of RMSE}

\begin{tabular}{lccccc}
\hline
\multirow{2}{*}{Regions}          & Metrics & \multicolumn{2}{c}{RMSE (MW)}     & \multicolumn{2}{c}{MAPE (\%)}\\ & \diagbox[width=1cm]{S}{E}     & N               & Y               & N             & Y             \\ \hline
\multirow{2}{*}{East Midlands}    & N       & 242.72         & \textbf{216.33}          & 8.62          & \textbf{7.72}          \\
                                  & Y       & 228.07          & 222.97 & 8.20          & 7.85 \\
\multirow{2}{*}{West Midlands}    & N       & 115.31          & 109.53          & 3.53          & 3.34          \\
                                  & Y       & 109.05          & \textbf{107.91} & 3.32          & \textbf{3.27} \\
\multirow{2}{*}{South Wales}      & N       & 103.11          & 109.18          & 8.46          & 9.00          \\
                                  & Y       & \textbf{101.82} & 108.61          & \textbf{8.35} & 8.93          \\
\multirow{2}{*}{South West}       & N       & 124.23          & 117.25          & 9.20          & 8.83          \\
                                  & Y       & 116.69          & \textbf{116.15} & 8.78          & \textbf{8.77} \\
\multirow{2}{*}{Northern Ireland} & N       & 53.66           & 49.42           & 5.64          & 5.15          \\
                                  & Y       & 50.03           & \textbf{48.68}  & 5.28          & \textbf{5.07} \\
\multirow{2}{*}{Ireland}          & N       & 154.08          & 165.90          & 3.52          & 3.88          \\
                                  & Y       & \textbf{149.54} & 160.16          & \textbf{3.46} & 3.76          \\ \hline
\end{tabular}
\end{table}

Table~\ref{Model performance table of RMSE} suggests that in each region, incorporating social features enhances forecasting accuracy. 
In almost all regions, economic indicators also contribute to better predictions.
The overlap effect of economic and social factors can be observed in the West Midlands, South West, and Northern Ireland.
For example, in the West Midlands, economic and social factors can improve the performance of the benchmark model by 5.0\% and 5.4\% respectively. 
The improvement can reach up to 6.4\% if both factors are combined.
According to the performance of the factors, we summarised the economic-sensitive regions as the East Midlands and Northern Ireland, and the rest are social-sensitive regions.

\begin{figure}[ht]
\centering
\includegraphics[scale=.35]{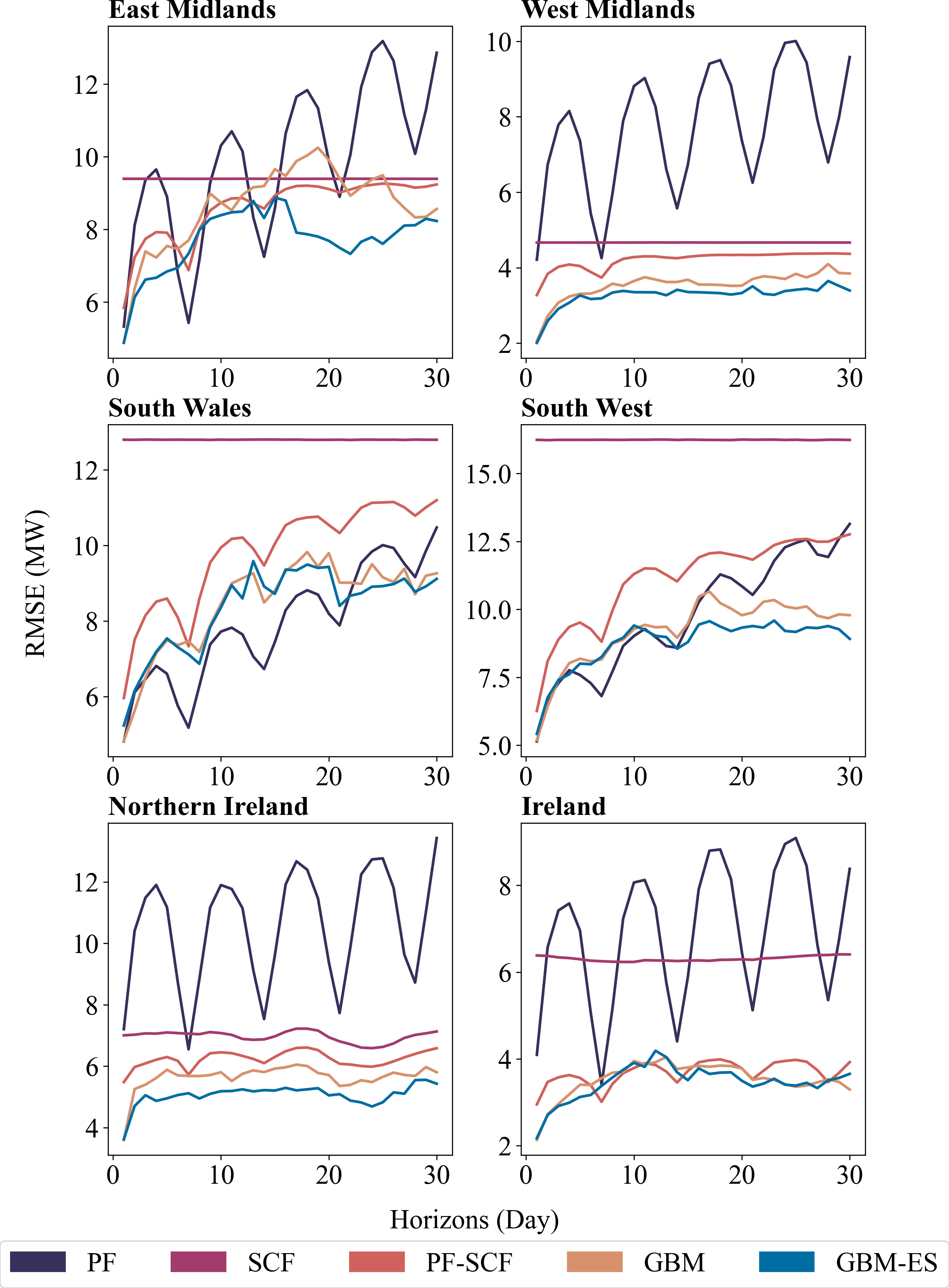}
\caption{MAPE plots for model PF, SCF, PF-SCF, GBM, and GBM-ES in all the regions.}
\label{RMSEBenchmarkPlots}
\end{figure}

The evolution of the forecasting error along the horizons from 1 to 30 days is presented in Figure~\ref{RMSEBenchmarkPlots}. 
The models GBM and GBM-ES are compared to naive forecasts: PF, SCF, and PF-SCF described in Appendix~\ref{Overview of Benchmark Models}.
The charts show how, in almost all cases, the proposed model outperforms naive approaches, and the use of social and economic factors reduces the errors.
These factors contribute more significantly to improving the accuracy in medium- and long-term forecasts.

\subsection{Probabilistic Forecast} \label{Results: Probabilistic forecast}

\deleted{
Before applying the PGBM model, it is essential to verify whether electricity demand follows a Gaussian distribution. 
We evaluate normality using the Kolmogorov-Smirnov (K-S) test and Z-score analysis. 
The K-S test compares the empirical distribution with a theoretical Gaussian distribution, while the Z-score method evaluates whether data points fall within three standard deviations ($\pm 3\sigma$) of the mean, as expected under a Gaussian distribution.
These tests are applied to data in each hour across multiple regions, with the K-S test providing a statistical measure of normality and the Z-score checking for extreme deviations. 
Table shows the proportion of hours in each region that pass each normality test.}

\deleted{
In Table~\ref{normality_results}, more than half of the 24 hours pass the K-S test in four out of six regions, suggesting that electricity demand follows a Gaussian distribution in most cases.
The Z-score test further confirms that a significant portion of the data remains within the expected range of $\pm 3\sigma$.
For consistency and fair comparison across regions, we assume that electricity demand follows a Gaussian distribution, and we use the PGBM model to estimate the mean and variance. 
Some hours in certain regions do not fully satisfy the Gaussian assumption, but exploring their exact distribution is beyond the scope of this study.}

\subsubsection{\added{Model Introduction}}
\added{
Traditional quantile-based probabilistic forecasting can be computationally expensive.
This study employs PGBM \cite{sprangers2021probabilistic}, which primarily learns the parameters of an assumed parametric conditional distribution for the target variable, typically as functions of input features.
These learned parameters allow for the reconstruction of the full predictive distribution, from which desired quantiles and prediction intervals can then be extracted.}

\added{
For parametric density forecasting of electricity demand, the Gaussian (or Normal) distribution is a common choice, due to its mathematical simplicity and ease of interpretation \cite{hong2016probabilistic}. 
Accordingly, this study assumes a conditional Gaussian distribution for electricity demand $y$ given input features $x$, denoted as: $y|x \sim N(\mu_y(x),\sigma_y^2(x))$.
The PGBM is trained to learn the mean $\mu_y(x)$ and variance $\sigma_y^2(x)$.
We acknowledge this as a simplification, as true conditional demand distributions may exhibit more complex characteristics.
The appropriateness of this assumption and the calibration of the resulting probabilistic forecasts are therefore evaluated in Section~\ref{Conditional Gaussian Diagnostics}.
}

\subsubsection{Forecasting Results and Improvements}

The overall CRPS values for each region are presented in Table~\ref{CRPS results on different models and horizons} to show the forecasting ability of social and economic factors.
The CRPS values are averaged on 30 horizons.
Table~\ref{CRPS results on different models and horizons} shows that CRPS can be decreased by 5.9\% on average by adding social factors in the regions except Ireland.
Economic factors improve probabilistic forecasts by 8.1\% on average in East Midlands, West Midlands, South West, and Northern Ireland.

\begin{table}[ht]
\centering
\caption{Horizon-averaged model performance of CRPS with and without economic and social factors. S and E indicate social and economic factors. N (No) and Y (Yes) are indicators to show if the factors (S or E) participate in the model.}
\label{CRPS results on different models and horizons}
\begin{tabular}{lccc}
\hline
Regions          & \diagbox{S}{E}          & N       & Y          \\ \hline
\multirow{2}{*}{East Midlands}    & N & 200.05          & 171.43          \\
                 & Y    & 185.60          & \textbf{170.70} \\
\multirow{2}{*}{West Midlands}    & N & 83.54          & 82.05          \\
                 & Y    & \textbf{75.87} & 77.70          \\ 
\multirow{2}{*}{South Wales}      & N & 80.41          & 89.11          \\
                 & Y    & \textbf{79.45} & 87.53          \\ 
\multirow{2}{*}{South West}       & N & 89.96          & \textbf{84.80} \\
                 & Y    & 85.15          & 85.13          \\ 
\multirow{2}{*}{Northern Ireland} & N & 36.56          & \textbf{32.74} \\
                 & Y    & 34.11          & 32.36          \\ 
\multirow{2}{*}{Ireland}          & N & \textbf{106.28}          & 123.80          \\
                 & Y    & 111.37 & 125.10          \\ \hline
\end{tabular}
\end{table}

\begin{figure}[ht]
\centering
\includegraphics[scale=.25]{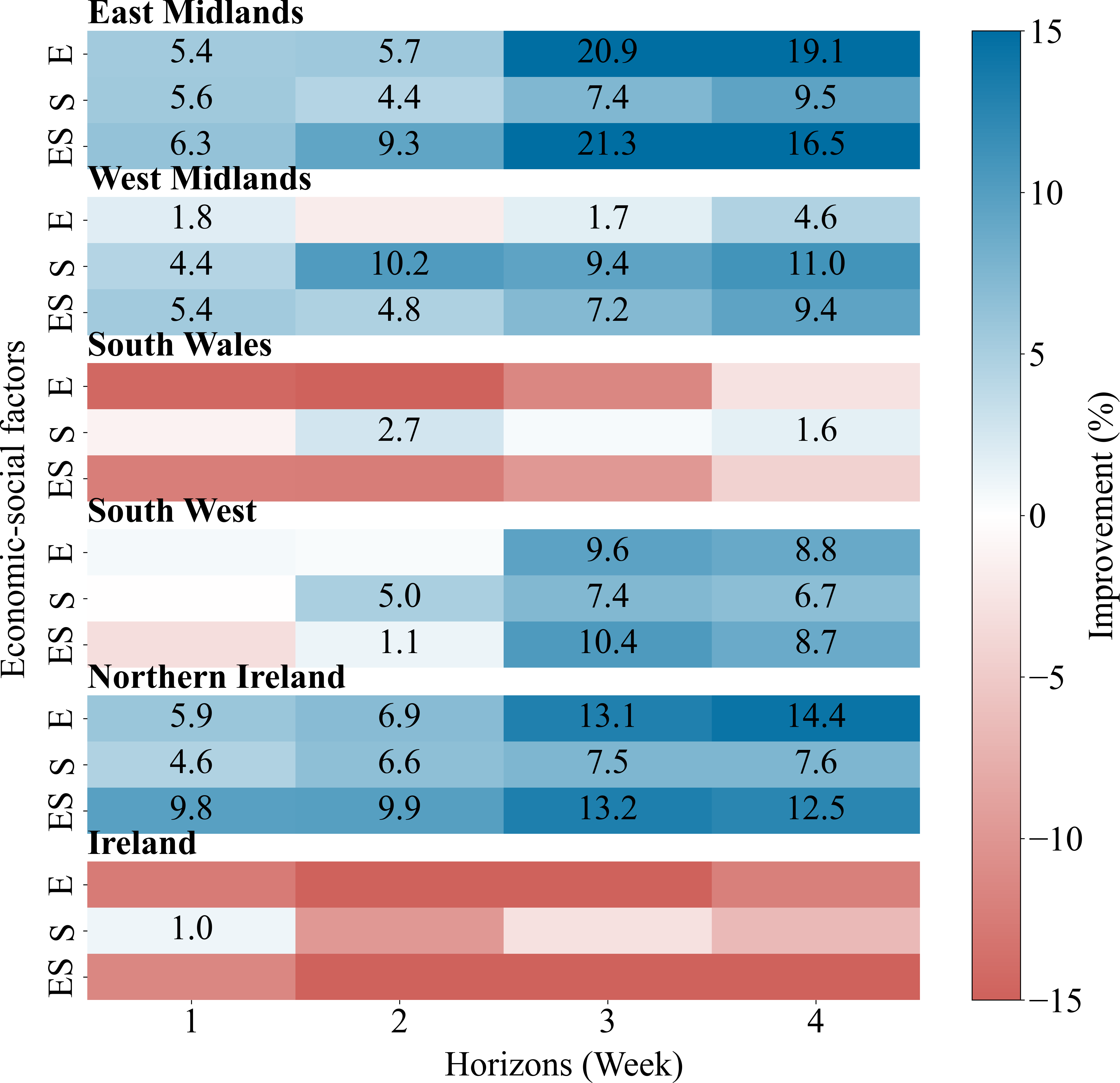}
\caption{Probabilistic forecasting improvements of models with economic and social factors. `E' for economic, `S' for social, `ES' for both economic and social. Only improvements over 1\% are shown in the blue grids.}
\label{ImpCRPS_week}
\end{figure}

Furthermore, we plot the CRPS improvements by percentage on all the horizons that are grouped by 4 weeks in Figure~\ref{ImpCRPS_week}.
Figure~\ref{ImpCRPS_week} demonstrates that in the East Midlands, economic and social factors improve forecasting all horizons, especially in the third and fourth weeks.
In West Midlands, social factors bring 4.4\% improvement for CRPS in the first week.
The improvement is growing in the following weeks and reaches 11\% at the last week.
Social and economic factors can both improve CRPS in South West and Northern Ireland, but higher improvements are observed after two weeks for South West.
The additive effect of social and economic features can be found at the first two weeks in Northern Ireland due to higher improvements from both external features compared to any single source.
In South Wales, only social features can enhance forecasting in the second and fourth week.
The incorporation of external features in Ireland has resulted in minimal observable improvement.

\subsubsection{\added{Conditional Gaussian Diagnostics}}\label{Conditional Gaussian Diagnostics}

\deleted{Figure~\ref{Reliability diagrams for East Midlands at Horizons 1 and 30 days.} presents reliability diagrams for the East Midlands, comparing the probabilistic calibration of forecasts with and without text features.
The results suggest that for most probability bins, predictions closely follow the perfect calibration line, indicating well-calibrated probabilistic forecasts. 
However, at lower probability levels (near 0.1), both models slightly underestimate the actual observed frequency. 
Moreover, the inclusion of text features does not significantly change calibration performance.}

\added{
This subsection provides an ex-post diagnostic evaluation of the PGBM's conditional Gaussian assumption, which is key to understanding its probabilistic behaviour and uncertainty estimate reliability.
We use two graphical tools: Quantile-Quantile (Q-Q) plots visually assess the normality of standardised residuals, focusing on tail behaviour to check the error distribution shape.
Reliability diagrams, using the Probability Integral Transform (PIT), evaluate overall forecast calibration by testing the uniformity of the PIT values' empirical cumulative distribution function (ECDF) against the ideal diagonal.}

\begin{figure}[ht]
\centering
\includegraphics[scale=.36]{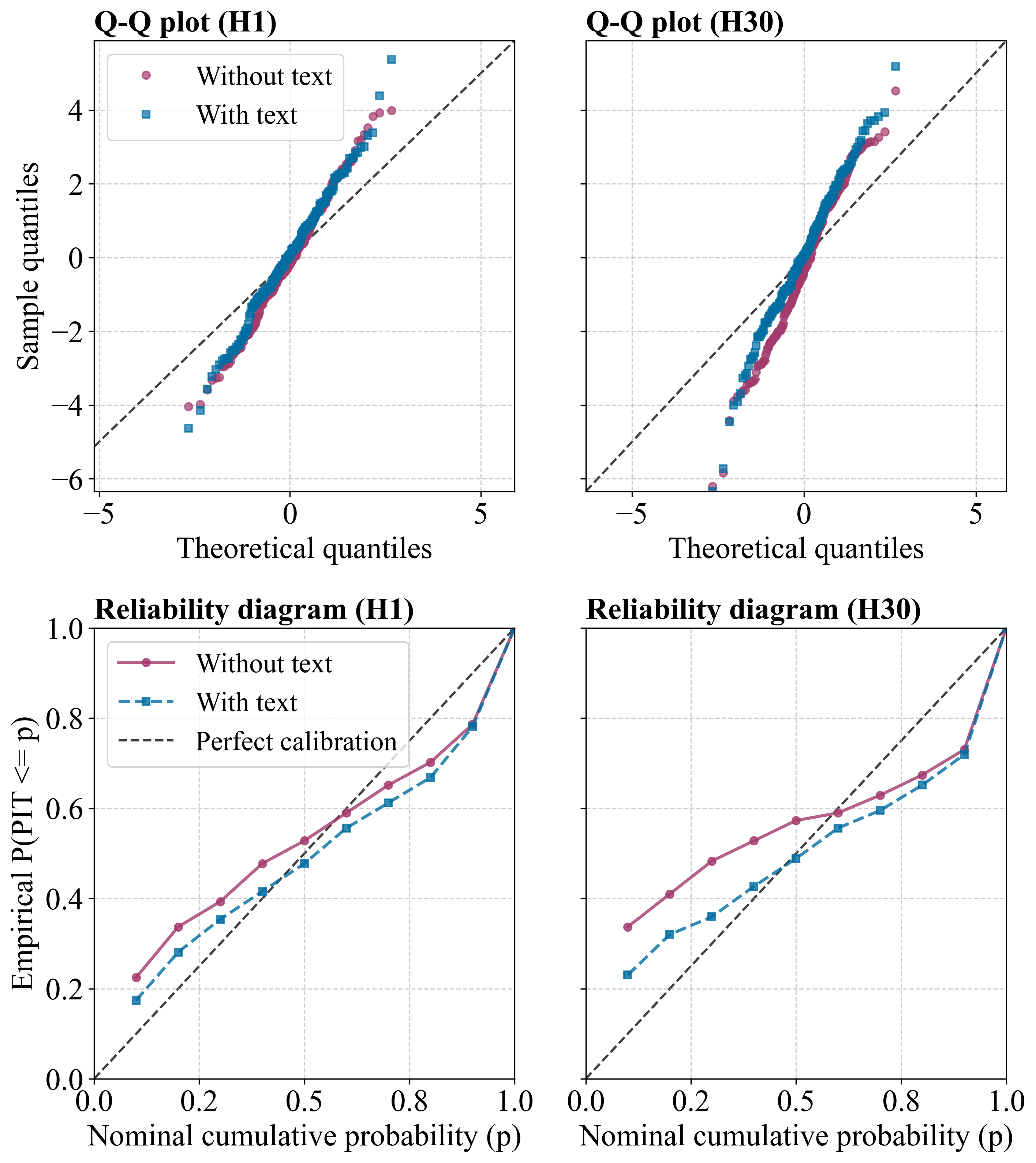}
\caption{Q-Q plots and reliability diagrams for West Midlands  (Hour 20, H1 and H30): models with and without text.}
\label{Diagnostic_Compare_H1H30_WestMidlands_Hr20}
\end{figure}

\added{Presenting exhaustive diagnostics is impractical due to the multi-layered data (regions, horizons, hours). 
Thus, for illustration, we analyse the West Midlands region (where text features notably improved CRPS, see Figure~\ref{ImpCRPS_week}) at Hour 20 (an evening peak).
We compare models with and without textual features for 1-day (H1) and 30-day (H30) ahead forecasts, corresponding to one-day and one-month lead times.
Figure~\ref{Diagnostic_Compare_H1H30_WestMidlands_Hr20} displays these diagnostics. 
The upper panels are Q-Q plots (theoretical vs. sample quantiles of standardised residuals) for H1 and H30, where the diagonal signifies perfect normality.
The lower panels show reliability diagrams (nominal cumulative probability $p$ vs. empirical P($\text{PIT} \leq p$)), where the diagonal indicates perfect calibration.}

\added{
In the Q-Q plots, all forecasts show reasonable central alignment with normality but deviate in the tails. 
Both H1 and H30 exhibit heavier lower tails with sample quantiles below the diagonal, and heavier upper tails with sample quantiles above the diagonal, when compared to a Gaussian distribution. 
The reliability diagrams for all forecasts show a bow shape, starting above the diagonal for low $p$ and falling below it for high $p$, a pattern indicative of underdispersion. 
For both Q-Q and reliability plots, the H1 forecasts, and those incorporating text features, generally appear closer to the ideal diagonal lines.
These ex-post diagnostics suggest that while the conditional Gaussian assumption is a simplification, particularly evident in the tail behaviour and overall underdispersion, the PGBM framework demonstrates relative improvements with text features and for the shorter H1 horizon in these visualisations. 
Although the Q-Q plot shows that the real distribution diverges from the Gaussian, the Reliability diagram shows a reduced divergence with the use of textual features, supporting the main hypothesis of this work.
}

\subsubsection{Impact of Text on Mean and Standard Deviation}

We conducted a paired t-test to assess whether textual features significantly impact the predicted mean ($\mu$) and standard deviation ($\sigma$).
For each forecast horizon and hour, predictions with and without text features were compared. 
The null hypothesis ($H_0$) assumes no difference, while the alternative hypothesis ($H_1$) suggests a significant change in $\mu$ or $\sigma$. 
The test results quantify whether text features affect the forecast central tendency of the distribution and dispersion across regions.
The results are presented in Table~\ref{Impact of Text Features on mean and standard deviation}.
 
\begin{table}[ht]
    \centering
    \caption{Impact of text on $\mu$ and $\sigma$ (unit: MW). Values marked with ** indicate statistical significance at $p < 0.01$.}
    \label{Impact of Text Features on mean and standard deviation}
    \begin{tabular}{lcccc}
        \hline
        \multirow{2}{*}{Region} & \multicolumn{2}{c}{Without Economics} & \multicolumn{2}{c}{With Economics} \\
                                & $\Delta\mu$     & $\Delta\sigma$    & $\Delta\mu$   & $\Delta\sigma$   \\ \hline
        East Midlands           & -20.66**           & -6.15**          & 16.64**          & -2.31**         \\
        West Midlands           & -16.09**           & -4.82**          & 5.41**           & -1.67**         \\
        South Wales             & -12.53**           & -2.43**          & 0.13             & -0.77**         \\
        South West              & -5.73**            & -3.34**          & 1.74**           & -1.11**         \\
        Northern Ireland        & -3.95**            & -2.09**          & 5.06**           & -1.05**         \\
        Ireland                 & -4.7**             & -5.31**          & 2.77**           & -1.74**         \\ \hline
        \end{tabular}
\end{table}

Table~\ref{Impact of Text Features on mean and standard deviation} show that in the absence of economic features, textual features significantly reduce both $\mu$ and $\sigma$ across regions. 
However, when economic features are included, textual features increase $\mu$ while still reducing the $\sigma$. 
This suggests a complementary relationship between text and economic data: when economic data is available, text features supplement market expectations, enhancing forecast stability. 
Conversely, in the absence of economic data, text features illustrate a stronger influence, potentially dominating the prediction.

\subsection{Forecasting on Special Day}
\begin{figure}[ht]
\centering
\includegraphics[scale=.35]{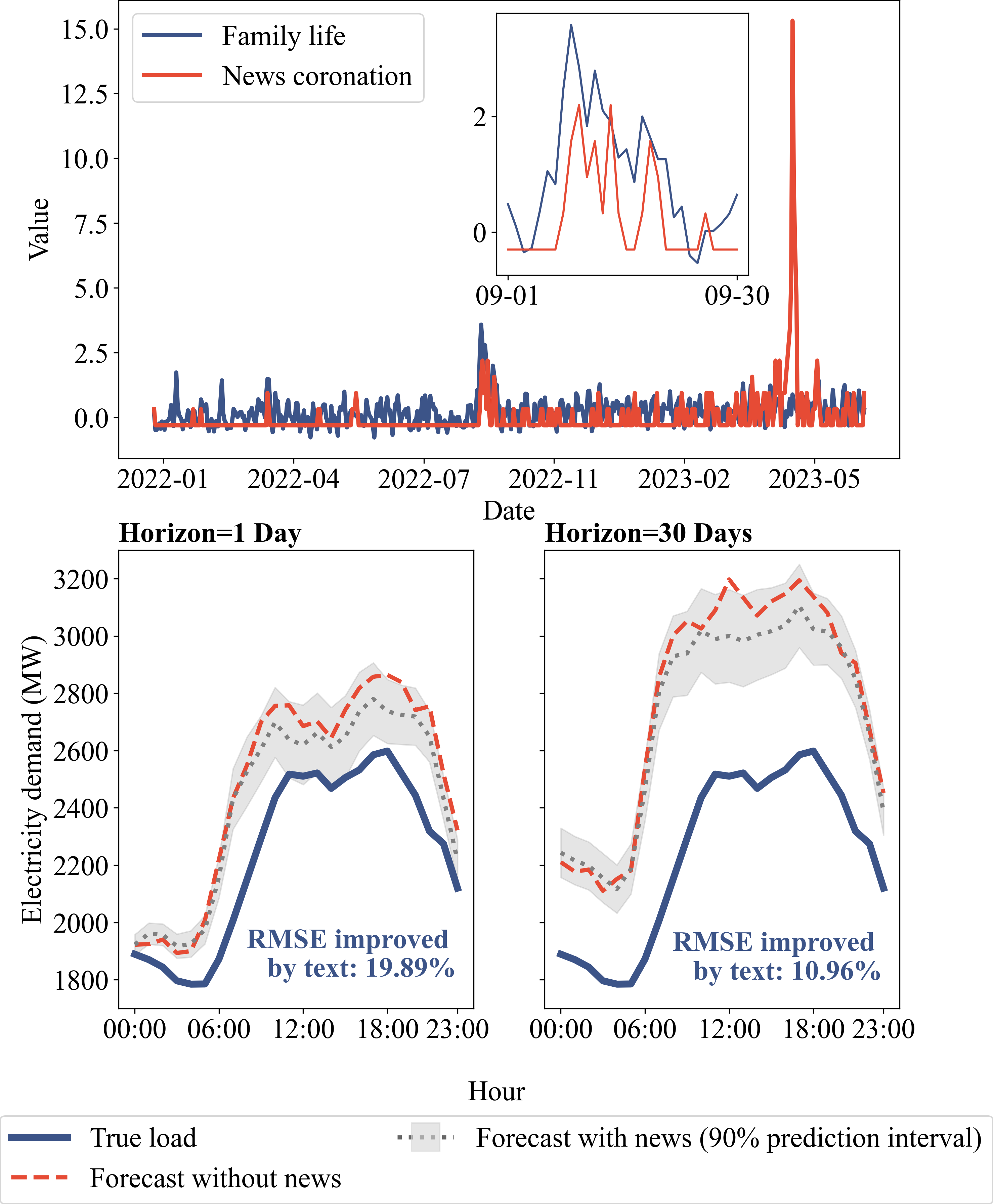}
\caption{\textbf{Top:} coronation-related news frequency (red line) and textual feature of family life (blue line). Both series are normalised.
\textbf{Bottom:} actual values and forecasts of electricity demand in the East Midlands on 2023-05-08. The forecasts are generated by benchmark and the model with social factors. The horizons are 1 and 30 days before 2023-05-08, which means that the datasets used in the two subplots end on 2023-05-07 and 2024-04-08 respectively. The RMSE improvements and 90\% prediction intervals are given in the figure.}
\label{Prob0508}
\end{figure}

Another analysis for forecasting was conducted on a single special day. 
In the UK, 2023-05-08 (Monday) was a bank holiday to celebrate King Charles III’s coronation. 
However, due to its exceptional nature, this date was not present in the calendar for identifying non-working days.
The government actually announced that 2023-05-08 would be a bank holiday on 2022-11-06, according to the news titled: \textit{Extra bank holiday approved to mark King's coronation} \cite{Extrabankholiday}.
The Python package \textsf{holidays 0.40} used in the benchmark model did not include 2023-05-08 as a holiday \cite{holidays-web}.
Thus, the benchmark model treated 2023-05-08 as a normal Monday and overestimated the demand on this day.
Nevertheless, the model incorporating social factors captured the relevant textual information that was mentioned more frequently as May 8th approached.
The model featuring social factors outperformed the benchmark on different horizons as shown in Figure~\ref{Prob0508}.

The top sub-plot of Figure~\ref{Prob0508} presents the frequency of news about the coronation ceremony, which is selected using the keywords \textit{king}, \textit{charles}, and \textit{coronation} from BBC news.
As May 8, 2023, approached, the media coverage and public interest in the coronation steadily intensified. 
The first peak in attention occurred in September 2022, while the second peak was observed in late April 2023.
The textual feature \textit{family life} used in our model, captured the first peak successfully as shown in the zoomed-in region of the top sub-plot. 
The term \textit{family life} also matches the British royal family.
The bottom sub-plots of Figure~\ref{Prob0508} show the improvement effects brought by this textual feature.
Although the demand is overestimated as shown by the red dashed line, the model with feature \textit{family life} can decrease forecasts especially in daytime, corresponding to the higher levels of human activity, and brings 19.89\% and 10.96\% of RMSE improvements for horizons 1 and 30 days.

This case illustrates the impact of sudden events on the benchmark model. 
The adjustments for holidays prevented the benchmark from responding in time, leading to forecasting biases. 
This demonstrates that our model with social factors has the ability to handle sudden events, especially when traces of these events can be found in historical news texts.

\subsection{Feature Importance Analysis} \label{Feature importance analysis}
\begin{figure*}[ht]
\centering
\includegraphics[scale=.55]{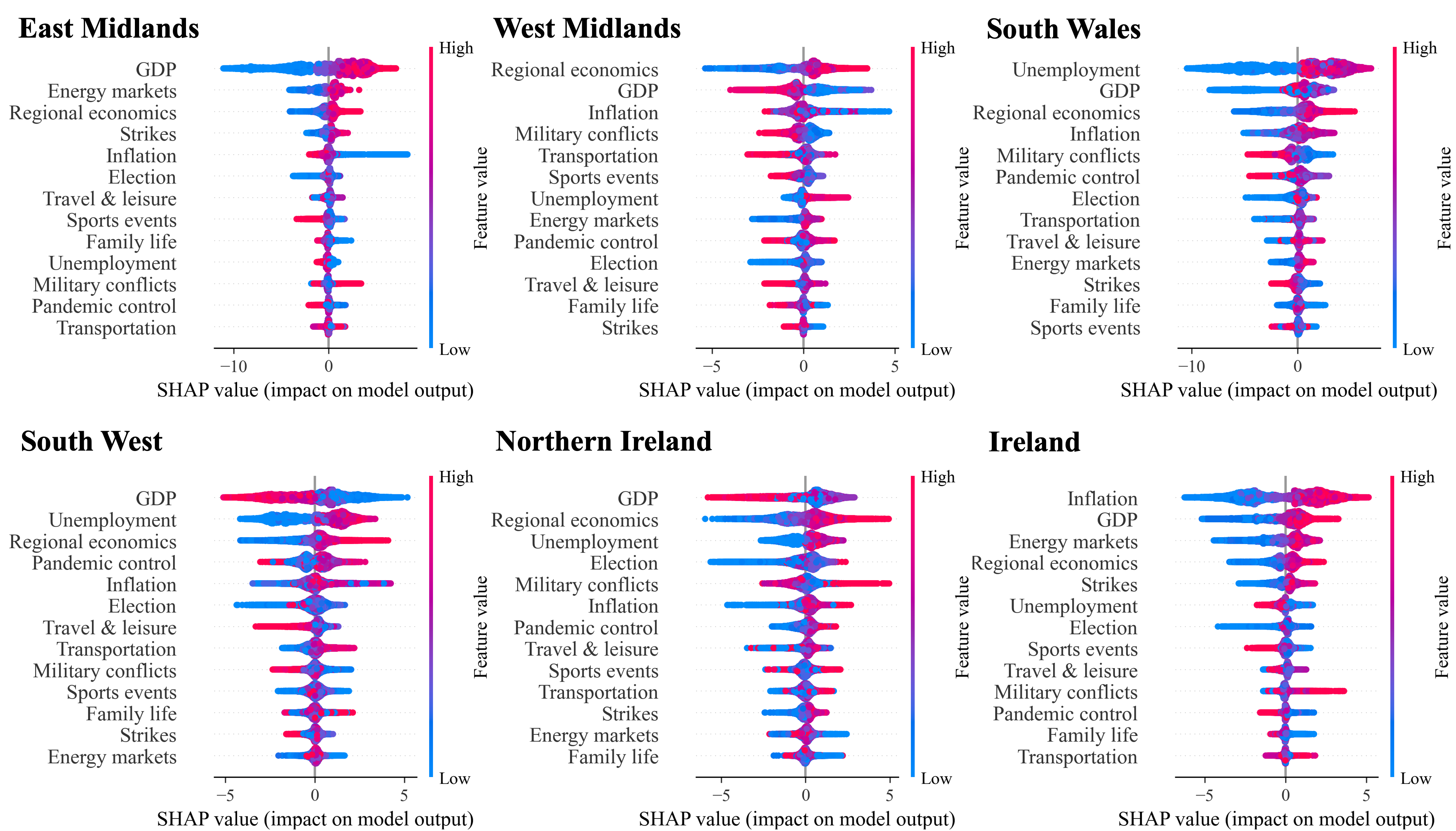}
\caption{Feature importance by SHAP values in all regions.}
\label{ShapBeesAll}
\end{figure*}

In this subsection, we applied SHAP (SHapley Additive exPlanations) method to explain how economic and social factors contribute to the forecasting process \cite{NIPS2017_8a20a862}.
We analysed the feature importance ranking for each region calculated by SHAP values as in Figure~\ref{ShapBeesAll}.
The SHAP values are averaged on all 30 horizons.
Figure~\ref{ShapBeesAll} demonstrates that GDP is a crucial factor for demand forecasting as it ranks first or second for all the regions.
\textit{Regional markets} is an important social factor, consistently ranking among the top three factors.

\begin{figure}[ht]
\centering
\includegraphics[scale=.7]{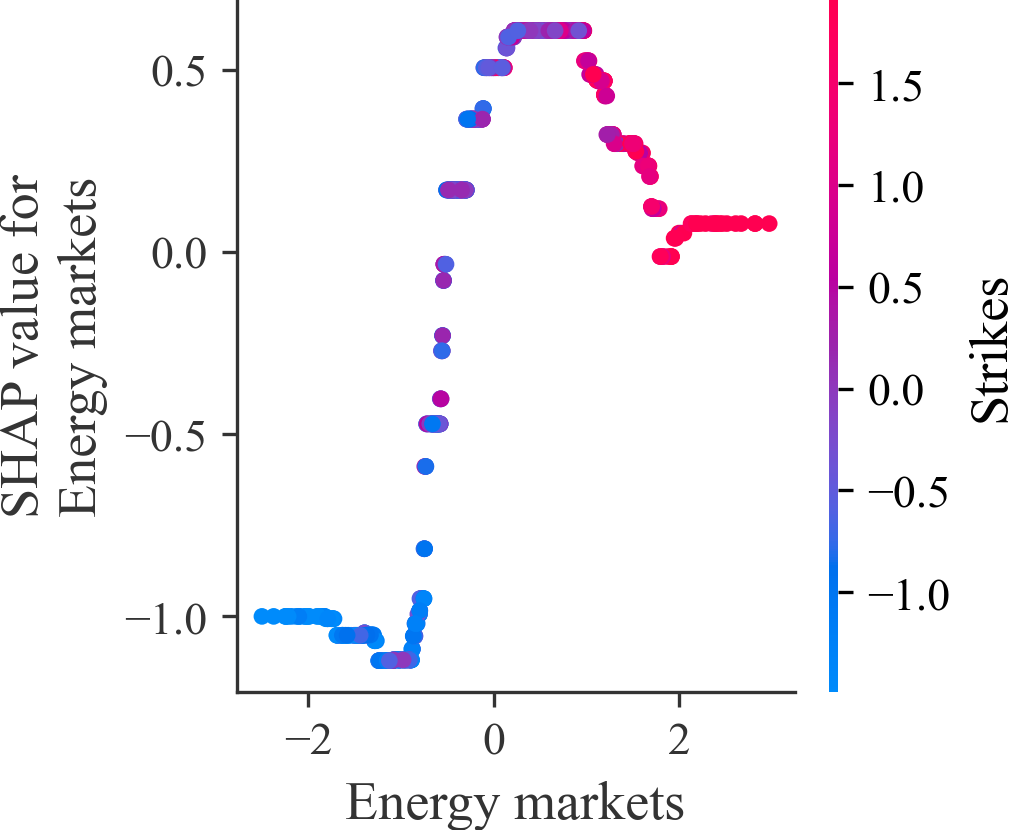}
\caption{Feature dependency of \textit{energy markets} and \textit{strikes} in the East Midlands.}
\label{ShapDependency}
\end{figure}

Figure~\ref{ShapDependency} illustrates the interaction between \textit{energy markets} and \textit{strikes}, detected by SHAP values for the East Midlands dataset. 
The SHAP values show an inverted U-shape.
At lower strike levels, SHAP values are positively correlated with energy markets, meaning higher mentions of energy markets contribute more to the predictions. 
However, as strike levels rise, SHAP values decline, suggesting a diminishing impact of energy markets under intensified strike conditions. 

To illustrate the contribution of individual features, we carried out an ablation experiment for each region.
This experiment examines the top features presented in Figure~\ref{ShapBeesAll} and sequentially incorporates each of them into the benchmark model.
Three economic features and two social features were selected. 
The RMSE and MAPE metrics are reported in Table~\ref{RMSE and MAPE results for single feature performance.}, where E1, E2, and E3 correspond to GDP, unemployment, and inflation.
$\text{S}_f$ and $\text{S}_j$ represent textual features of \textit{regional economics} and \textit{energy markets}.
The areas R1 to R6 are East Midlands, West Midlands, South Wales, South West, Northern Ireland, and Ireland.

\begin{table}[ht]
\centering
\caption{RMSE and MAPE results for single-feature performance. Bold and underlined values indicate the best and second-best features.}
\label{RMSE and MAPE results for single feature performance.}
\scalebox{0.88}{
\begin{tabular}{lccccccc}
\hline
Areas                           & Metrics & Benchmark & E1              & E2              & E3          & $\text{S}_f$     & $\text{S}_j$           \\ \hline
\multirow{2}{*}{R1}    & RMSE    & 242.72    & \textbf{224.12} & \underline {233.59}    & 238.97      & 242.71 & 241.17       \\
                                  & MAPE    & 8.62      & \textbf{8.02}   & \underline {8.26}      & 8.50        & 8.63   & 8.58         \\
\multirow{2}{*}{R2}    & RMSE    & 115.31    & \textbf{110.19} & \underline {110.91}    & 114.29      & 114.05 & 115.88       \\
                                  & MAPE    & 3.53      & \textbf{3.37}   & \underline {3.41}      & 3.52        & 3.49   & 3.57         \\
\multirow{2}{*}{R3}      & RMSE    & 103.11    & 106.95          & \textbf{100.40} & 108.34      & 103.20 & \underline {102.93} \\
                                  & MAPE    & 8.46      & 8.76            & \textbf{8.25}   & 9.00        & 8.49   & \underline {8.46}   \\
\multirow{2}{*}{R4}       & RMSE    & 124.23    & \textbf{111.11} & 122.26          & 133.67      & 123.64 & \underline {122.19} \\
                                  & MAPE    & 9.20      & \textbf{8.20}   & 9.11            & 10.24       & 9.16   & \underline {9.04}   \\
\multirow{2}{*}{R5} & RMSE    & 53.66     & \textbf{49.19}  & 53.38           & \underline {50.70} & 53.25  & 53.55        \\
                                  & MAPE    & 5.64      & \textbf{5.12}   & 5.61            & \underline {5.33}  & 5.61   & 5.65         \\
\multirow{2}{*}{R6}          & RMSE    & 154.08    & \underline {148.27}    & \textbf{141.77} & 169.99      & 153.78 & 149.03       \\
                                  & MAPE    & 3.52      & \underline {3.42}      & \textbf{3.20}   & 4.00        & 3.52   & 3.39         \\ \hline
\end{tabular}}
\end{table}

From Table~\ref{RMSE and MAPE results for single feature performance.}, GDP (E1) and unemployment (E2) are the two most influential features. 
Inflation (E3) enhances demand forecasting in the East Midlands, West Midlands, and Northern Ireland. 
Both regional economics ($\text{S}_f$) and energy markets ($\text{S}_j$) contribute to improved forecasting in five out of the six regions. 
However, when considering individual features, economic factors, particularly GDP and unemployment, play a more significant role in improving forecasting accuracy.

\subsection{Causality Analysis} \label{Causality analysis}

This study aims not only to utilise textual information to enhance forecasting but also to explore the relationship between demand and social factors.
This subsection presents a causality analysis between textual features and regional demand, based on results from the Granger and DML tests.
The Granger test evaluates whether incorporating a feature into the forecasting model improves accuracy, and it is applied to each feature and demand variable across all regions.
Table~\ref{Granger causality of textual features and regional demand.} presents the results of the Granger test. 
In the table, the regions R1 to R6 are East Midlands, West Midlands, South Wales, South West, Northern Ireland, and Ireland.
A $\rightarrow$ indicates that the feature in the row Granger-causes demand in the corresponding column region, a $\leftrightarrow$ is bidirectional causality between the feature and demand, and a $\leftarrow$ denotes that demand in the column region Granger-causes the feature in the row.

\begin{table}[ht]
\centering
\caption{Granger causality of text and regional demand.}
\label{Granger causality of textual features and regional demand.}
\begin{tabular}{lcccccc}
\hline
Textual features & R1     & R2     & R3       & R4        & R5  & R6           \\ \hline
Military conflicts  & $\rightarrow$     & $\rightarrow$     & $\rightarrow$     & $\rightarrow$     & $\leftrightarrow$ & $\leftrightarrow$ \\
Transportation           & $\rightarrow$     & $\rightarrow$     & $\rightarrow$     & $\rightarrow$     & $\leftrightarrow$ & $\rightarrow$     \\
Travel \& leisure           & $\leftarrow$      &                   &                   &                   & $\leftrightarrow$ & $\rightarrow$     \\
Sports events           &                   &                   &                   &                   &                   &                   \\
Pandemic control            & $\leftrightarrow$ & $\leftrightarrow$ & $\leftrightarrow$ & $\leftrightarrow$ & $\leftrightarrow$ & $\leftrightarrow$ \\
Regional economics            & $\leftrightarrow$ & $\leftrightarrow$ & $\rightarrow$     & $\leftrightarrow$ & $\leftrightarrow$ & $\leftrightarrow$ \\
Strikes            & $\rightarrow$     & $\leftrightarrow$ & $\leftrightarrow$ & $\leftrightarrow$ & $\leftrightarrow$ & $\leftrightarrow$ \\
Family life            &                   &                   & $\leftarrow$      & $\rightarrow$     & $\leftrightarrow$ & $\leftrightarrow$ \\
Election           &                   &                   &                   & $\rightarrow$     & $\leftrightarrow$ & $\rightarrow$     \\
Energy markets            & $\rightarrow$     & $\rightarrow$     & $\rightarrow$     & $\leftrightarrow$ & $\leftrightarrow$ & $\leftrightarrow$ \\ \hline
\end{tabular}
\end{table}

Table~\ref{Granger causality of textual features and regional demand.} shows that events related to military conflicts, transportation, and energy markets show Granger causality with electricity demand across all regions.
These events are also Granger-caused by demand in Northern Ireland.
Bidirectional Granger causality is observed for pandemic control, regional economics, and strikes. 
Notably, the feature representing sporting events does not exhibit a significant correlation in the Granger test. 
This is possibly due to the multi-horizon nature of the demand studied and the relatively weak long-term influence of such events.

\begin{figure*}[ht]
\centering
\includegraphics[scale=.4]{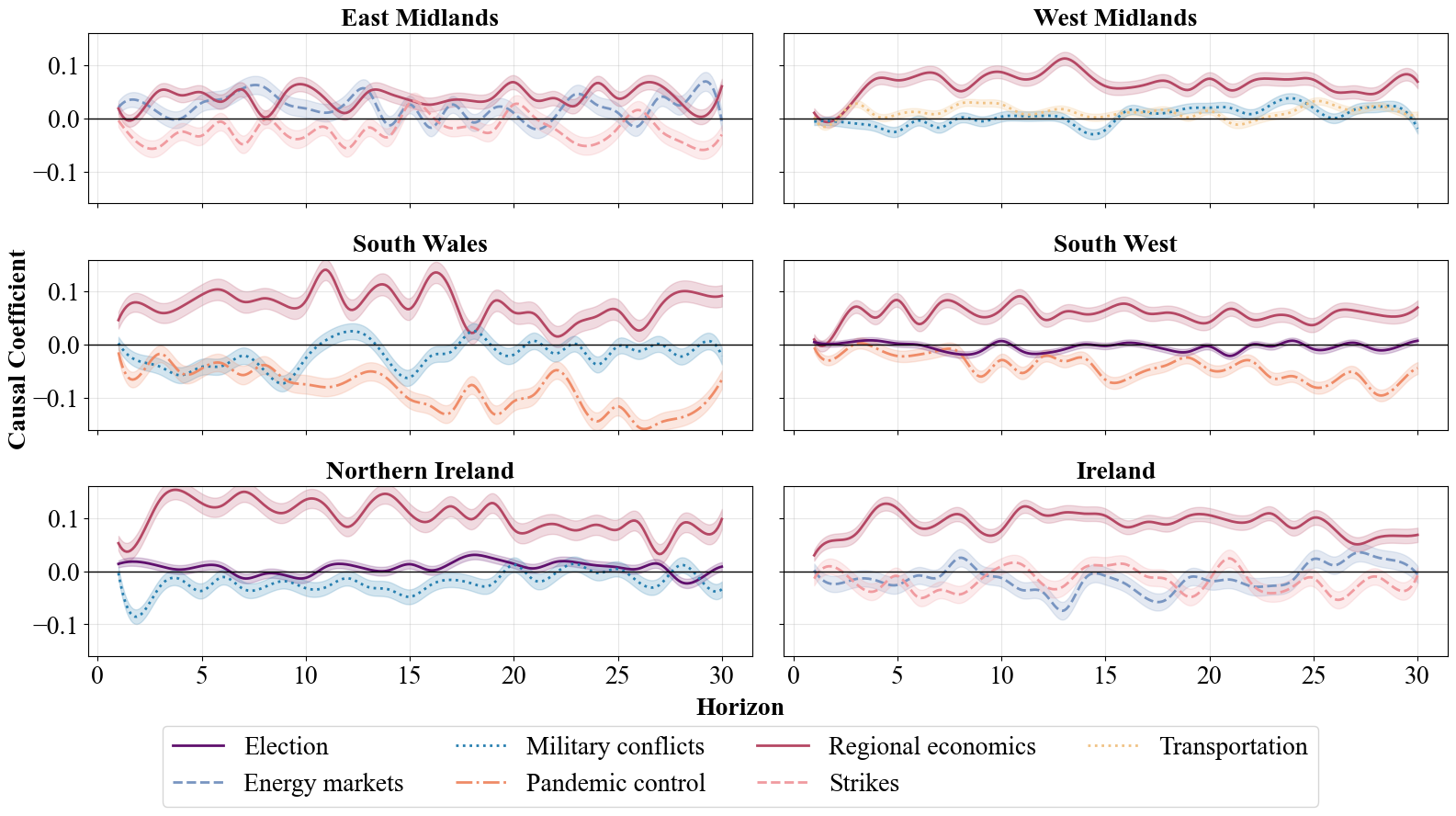}
\caption{Horizontal varying causal coefficients of textual features and demand in all areas.}
\label{Horizontal varying causal coefficients of textual features and demand in all areas.}
\end{figure*}

We then selected the three most influential textual features from Figure~\ref{ShapBeesAll} and conducted a DML test to isolate the causal effect of each feature on forecasting.
For a more detailed analysis, we calculated and plotted the causal coefficients across all 30 horizons, as shown in Figure~\ref{Horizontal varying causal coefficients of textual features and demand in all areas.}. 

The results indicate that regional economics positively influence electricity demand across all areas. 
The causal coefficient initially increases over the first five days before stabilising in subsequent horizons. 
Similarly, fluctuations in the energy market positively impact demand in the East Midlands, with SHAP values also showing a positive correlation with forecasts.
Conversely, several factors exhibit negative causality: strikes reduce demand in the East Midlands and Ireland, pandemic control decreases demand in South Wales and the South West, and military conflicts negatively impact demand in South Wales and Northern Ireland.

\section{Discussion}\label{Discussion}
From the analysis of the results above, this study aims at answering the following research questions, and is thus beneficial for a better understanding of electricity demand, its potential application in forecasting, and the advantages and limitations of using NLP. \\

\textbf{1. Is there any relationship between national news and regional electricity demand?}

This question was initially answered in \cite{bai2024newsandload} at the national level. 
This second study shows that there is still a stable relationship between national news and regional electricity demand.
Section~\ref{Causality analysis} identifies that social factors from national news, such as energy markets and regional economics, can result in a Granger-causal change in regional demand.
According to Section~\ref{Feature importance analysis}, news information plays a significant role in forecasting models and we can also detect the complex interaction between different economic or social factors. \\

\textbf{2. Can textual news be used for practical applications to improve regional demand forecasting?}

NLP techniques are powerful for extracting numeric features, such as sentiments and topics, from textual news, and have been used to forecast crude oil prices \cite{bai2022crude}, taxi demand \cite{rodrigues2019combining}, and so on.
In this study, adding textual features to our forecasting model improved both deterministic and probabilistic performance, as evidenced by Section~\ref{Results: Deterministic forecast} and \ref{Results: Probabilistic forecast}.
In Table~\ref{Model performance of RMSE and CRPS}, the results have been averaged across all regions and horizons.
They show better performance when textual features are added, with improvements of 3.9\% and 4.2\% for deterministic and probabilistic tasks, respectively.

\begin{table}[ht]
\centering
\caption{Horizon and regional averaged model performance of RMSE and CRPS with and without economic and social factors. S and E indicate social and economic factors. N (No) and Y (Yes) are indicators to show if the factors (S or E) participate in the model.}
\label{Model performance of RMSE and CRPS}
\begin{tabular}{lcccc}
\hline
\multirow{2}{*}{\diagbox{S}{E}} & \multicolumn{2}{l}{Deterministic-RMSE} & \multicolumn{2}{l}{Probabilistic-CRPS} \\
                  & N               & Y               & N               & Y               \\ \hline
N                  & 130.93          & 127.75          & 99.47        & 97.32         \\
Y                  & \textbf{125.78}         & 127.41          & \textbf{95.26}          & 96.42   \\ \hline
\end{tabular}
\end{table}

\textbf{3. Which topics have a higher impact on electricity demand?}

SHAP values offer a method to quantify feature importance in forecasting models.
As described in Section~\ref{Feature importance analysis}, the significance of features varies across regions.
Key economic indicators such as GDP, inflation, and unemployment are influential for regional demand. 
Additionally, social factors such as energy markets, regional economics, military conflicts, strikes, and transportation rank higher than other factors.\\

\textbf{4. Are the results consistent for different regions?}

The study reveals a mixed picture of improvements across the six regions under investigation. 
In the deterministic forecasting results, social factors enhance forecasting accuracy across all six regions, reducing RMSE by 6.0\%, 5.4\%, 1.3\%, 6.1\%, 6.8\%, 2.9\%.
Economic factors contribute to improved forecasting in the East Midlands, West Midlands, South West, and Northern Ireland, as shown in Table~\ref{Model performance table of RMSE}.
The ablation experiments on individual features (Table~\ref{RMSE and MAPE results for single feature performance.}) indicate that, in addition to economic factors, textual features related to regional economics and energy markets consistently improve the benchmark model in five regions.

In probabilistic forecasting, the results in Figure~\ref{ImpCRPS_week} show significant improvements in four regions by textual features. 
While the enhancements in South Wales and Ireland are marginal, textual features still contribute to improved deterministic forecasting in these regions.
In Ireland, this phenomenon was in part expected, since this dataset was added for comparison with the UK regions, but the model was fed with textual data from the UK, not from Ireland.
On the one side, Ireland is close to the UK, sharing a common border and one of its official languages, English.
Furthermore, news about politics in Northern Ireland was indicated as important for the prediction of UK national load in \cite{bai2024newsandload}. 
On the other hand, BBC news is clearly more related to the UK and not directly pertinent to Ireland, therefore, the lack of relevant impact observed is understandable.
Regarding South Wales, its demand has been found to be the most challenging to predict, as shown in Figure~\ref{RMSEBenchmarkPlots} and Figure~\ref{ImpCRPS_week}.
Since national-level UK news enhances regional demand forecasting, incorporating regional news could further optimise results. This will be explored in future work.\\

\textbf{5. For which forecast horizons is the impact higher?}

From Figure~\ref{ImpCRPS_week}, economic factors show higher improvements after the second week for the East Midlands, South West, and Northern Ireland.
Social factors show higher improvements after the first week for the West Midlands, South Wales, and Northern Ireland.\\

\phantomsection

\section{Conclusions}\label{Conclusions}
This study expands text-based forecasting applications in power systems by investigating the impact of national news on regional electricity demand.
We experimented on multi-horizon and multi-region electricity demand forecasting incorporating both economic and social factors.
The results of Granger causality tests indicate a causal relationship between national social factors and regional demand. 
Social factors extracted from news sources can enhance electricity forecasting, improving deterministic predictions by 3.9\% and probabilistic predictions by 4.2\% overall. 
Our analysis reveals that factors such as energy markets, regional economics, and military conflicts have a more profound impact on demand. 
In multi-horizon forecasts, social factors generally take effect after one week, aligning with the delay in the response of power systems to news events. 
We have also identified regions that are particularly sensitive to economic or social factors and provided targeted recommendations to help power system managers optimise their strategies.

\bibliographystyle{IEEEtran}
\bibliography{sn-bibliography}

\renewcommand{\thesubsection}{\Alph{subsection}}
\setcounter{subsection}{0}
\section*{Appendix: Benchmark Model Selection}
\subsection{Overview of Benchmark Models}
\label{Overview of Benchmark Models}
Three naïve models are built for multi-horizon forecasting: Persistence Forecasts (PF), Smart Climatology Forecasts (SCF), and a combination of PF and SCF (PF-SCF) \cite{murphy1992climatology}.
The PF model simply uses the last observed value \( y_t \) as the forecast for all horizons. 
The SCF model averages historical data for matching months and hours to generate forecasts.
Combining PF and SCF (denoted as PF-SCF) has been shown to outperform either model alone \cite{murphy1992climatology}. 
The combined forecast is expressed as a weighted sum of the two, \( \hat{y}_{PF-SCF} = \alpha \hat{y}_{PF} + (1-\alpha) \hat{y}_{SCF} \), where \( 0 < \alpha < 1 \). 
In this study, we use LASSO regression to determine the optimal weights.
These naïve models rely only on the time series, without any external information. 
As fundamental models, they establish a lower bound for forecasting performance.

Least Absolute Shrinkage and Selection Operator (LASSO) \cite{santosa1986linear} and Generalised Additive Models (GAM) \cite{obst2021adaptive} are considered in this study. 
Compared to the naïve models, these methods leverage external information, such as historical lags, calendar effects, and temperature. 
LASSO, as a linear model, captures linear relationships between these variables, while GAM provides flexibility by incorporating nonlinear components using techniques like splines.

Several machine models are applied: Support Vector Regression (SVR), ExtraTrees Regressor (ETR), and Light Gradient Boosting Machines (LGBM).
Different from the naïve and linear or semi-linear methods, these models can flexibly capture complex, nonlinear interactions in the data.
SVR uses a kernel-based method to map features into higher-dimensional spaces, making it suitable for modelling nonlinear relationships \cite{smola2004tutorial}. 
ETR is an ensemble-based tree method that partitions the feature space to model interactions and handles high-dimensional data efficiently \cite{geurts2006extremely}.
LGBM, a gradient-boosting framework, iteratively builds decision trees to minimise errors, leveraging both linear and nonlinear relationships among features \cite{ke2017lightgbm}. 
These models were included in our analysis to ensure robust feature utilisation and to select a flexible model that works well for the given prediction task.

\deleted{For probabilistic forecasting, GBM trains separate models for each target quantile to approximate the forecasting distribution. 
However, this approach can become computationally expensive when multiple quantiles are required. 
To address this, we adopt Probabilistic Gradient Boosting Machines (PGBM), which build upon the structure of GBM but focus on learning the parameters of an assumed target distribution instead of individual quantiles \cite{sprangers2021probabilistic}.  
We chose PGBM because LGBM has demonstrated strong performance in deterministic forecasting tasks, as in Figure~\ref{FriedmanNemenyi}, making it a natural candidate to extend to probabilistic forecasting scenarios.} 

\deleted{In this case, we assume that the target data follows a Gaussian distribution, enabling PGBM to directly learn the distribution parameters, such that \( (\mu_{\hat{y}}, \sigma_{\hat{y}}^2) = f(\bm{x}) \), where \( \mu_{\hat{y}} \) and \( \sigma_{\hat{y}}^2 \) are the predicted mean and variance, respectively. 
The forecasted target \( \hat{y} \) can then be sampled or inferred from the Gaussian distribution \( \hat{y} \sim \text{N}(\mu_{\hat{y}}, \sigma_{\hat{y}}^2) \). 
This approach not only simplifies the training process but also effectively models the forecasting uncertainty in scenarios where the Gaussian assumption holds.}

\subsection{Forecasting Results of Benchmark Models}\label{Forecasting Results of Benchmark Models}

This subsection provides the results comparison of the candidate benchmarks.
Considering there are six regions and 30 forecasting horizons in each region, it is redundant to demonstrate every single result. 
Instead, we treat each horizon in each region as a distinct forecasting task and compare only the model rankings on the 180 tasks for simplicity. 
We performed Friedman and Nemenyi tests to examine if there are significant differences in model rankings across all tasks \cite{demvsar2006statistical}.

\begin{figure}[ht]
\centering
\includegraphics[scale=0.27]{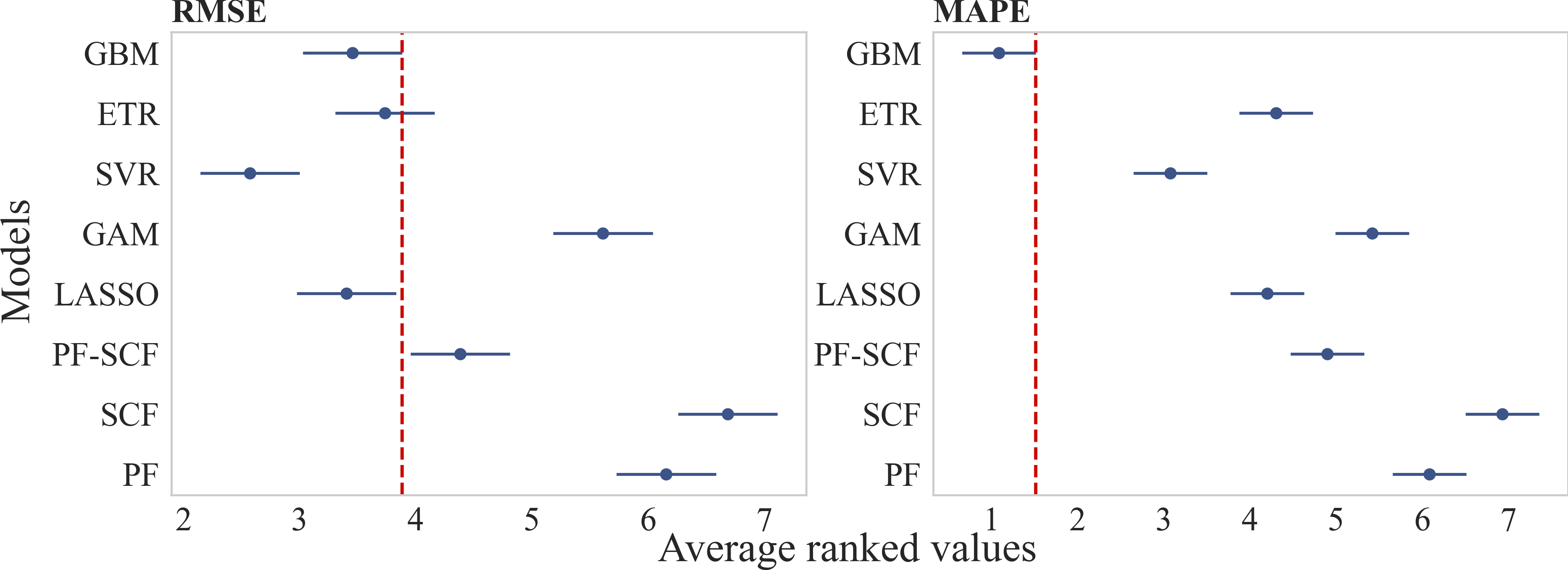}
\caption{Friedman and Nemenyi test results of benchmark models. \textbf{Left}: RMSE ranking. \textbf{Right}: MAPE ranking.}
\label{FriedmanNemenyi}
\end{figure}

In Figure~\ref{FriedmanNemenyi}, each point represents the average rank of a model on the 180 forecasting tasks.
The short lines through the points are thresholds given by Nemenyi test.
If the rank value of model A is lower than model B, and there is no overlap of the short lines, then model A ranks significantly higher than model B in all the tasks.
We added a red dashed line to the right end of the GBM interval to facilitate comparison.
From the left panel of Figure~\ref{FriedmanNemenyi}, SVR is the best-performing model, and there is no significant difference in the ranks of GBM, ETR, and LASSO.
The four models are all better than GAM and naive models.
In the MAPE results, GBM ranks significantly first among the models.

\begin{IEEEbiography}[{\includegraphics[width=1in,height=1.25in,clip,keepaspectratio]{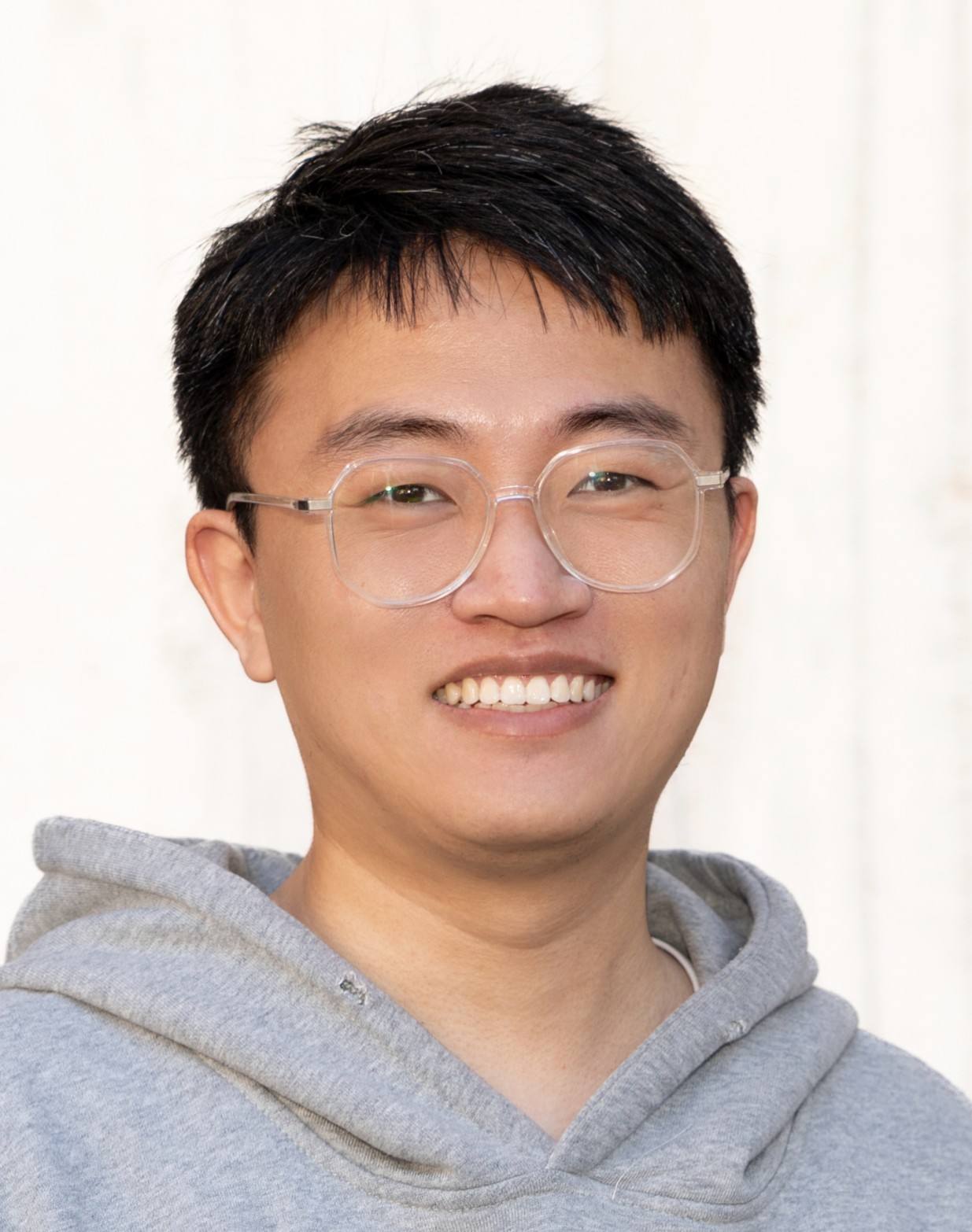}}]{Yun Bai} (Member, IEEE) received a BSc degree in Statistics and an MSc degree in Management Science and Engineering from Beihang University, Beijing, China.
Since 2022, he has been a doctoral student at MINES Paris - PSL University with a scholarship from the China Scholarship Council (CSC). His research interests include energy forecasting, natural language processing applications, and machine learning techniques.
\end{IEEEbiography}
\vspace{-10mm}

\begin{IEEEbiography}[{\includegraphics[width=1in,height=1.25in,clip,keepaspectratio]{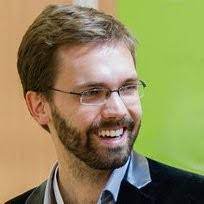}}]{Simon Camal} (Member, IEEE) received a BEng. degree in Energy and Environmental Engineering from Mines Nancy, Nancy, France, in 2010, a European MSc in Renewable Energy from Loughborough University, Loughborough, UK, in 2011, and a PhD from MINES Paris - PSL University, Paris, France, in 2020, on forecasting and optimization of ancillary service provision by renewable energy power plants. He currently works at the MINES ParisTech Center for Processes, Renewable Energies and Energy Systems (PERSEE), Sophia Antipolis, France, as the Project Manager of the Horizon2020 Smart4RES Project.
\end{IEEEbiography}
\vspace{-10mm}

\begin{IEEEbiography}[{\includegraphics[width=1in,height=1.25in,clip,keepaspectratio]{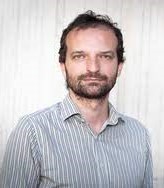}}]{Andrea Michiorri} MSc, PhD, HDR, is a Researcher at the PERSEE Centre of Mines Paris – PSL where he leads research on the integration of renewables into the power systems. 
\end{IEEEbiography}
\end{document}